# Reading a Legal Question Word by Word: Embedding Trajectories of 2,144 Vietnamese Legal Headlines

Tran Minh Quan

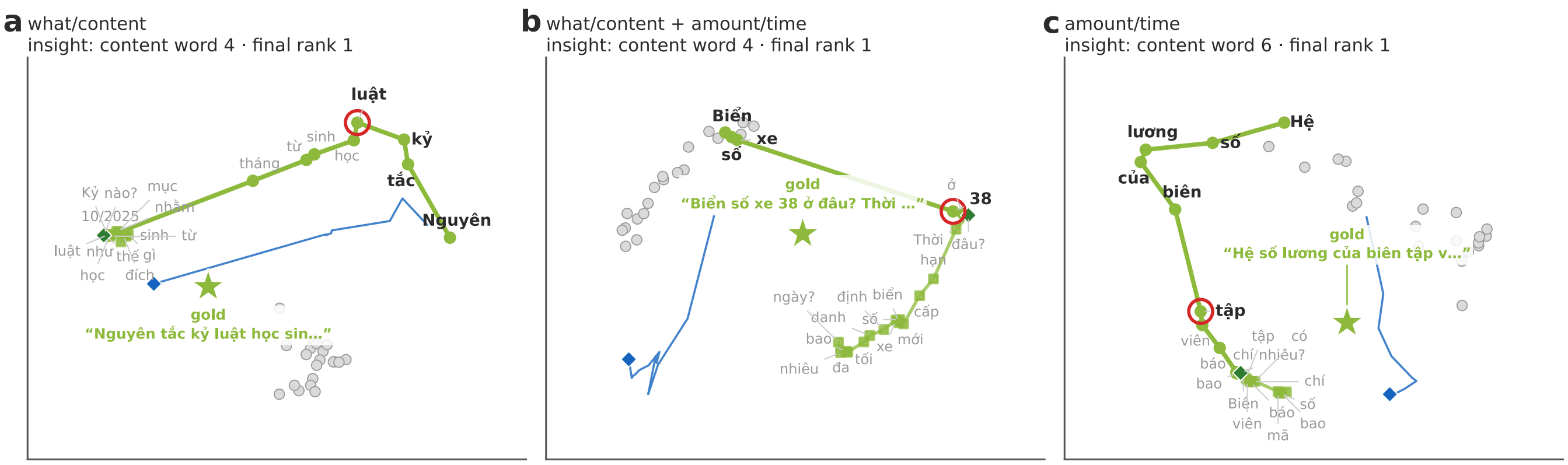

Figure 1: **Three headlines walking to their answers, one word at a time** (Nemotron-3-Embed-8B). Each panel is the plane of the gold article (**green star**, labelled *gold* with its opening words in green), its twenty strongest competitors (**grey dots**) and the word-prefix path of the headline (**green line**: one **green circle** per word of the first sub-question, **green squares** for the words of the second). Every word is written where the prefix ending in it is encoded, joined to its point by a **light-grey leader**: **bold black words** are the content words needed to bring the gold article to rank 1, up to the **red ring** (the *insight point*); **grey words**—the interrogative frame, the greeting and the attribution—move the point but not the ranking, and **grey dots** at the start of a path are greeting words. The **thin blue line** ending in a **blue diamond** is the second sub-question encoded on its own; it lands elsewhere. (a) *Nguyên tắc kỷ luật học sinh ...* (Principles for disciplining pupils ...): rank 1 after four content words. (b) *Biển số xe 38 ở đâu?* (Where is licence plate 38 from?): four words, the number does the work. (c) *Hệ số lương của biên tập viên ...* (Salary coefficient of editors ...): six words.

## ABSTRACT

A dense retriever encodes a question as one vector, but the question arrives one word at a time. We read every one of 2,144 held-out headlines from the Vietnamese legal library *Thư Viện Pháp Luật* (Legal Library) word by word through four decoder embedders—NVIDIA's Nemotron-3-Embed-8B and Nemotron-3-Embed-1B, Alibaba's Qwen3-Embedding-8B and Qwen3-Embedding-0.6B—encoding all 65,444 word prefixes, ranking each against the 20,034-article corpus, and keeping the full vector of every prefix. We then split the 1,112 multi-question headlines into their 3,438 sub-questions, encode every prefix of every sub-question on its own, and do the same for 168 answers read word by word. The trajectories show one behaviour with a few well-defined variants. (i) The gold article becomes rank 1 after a median of 6–7 content words in every encoder (quartiles 4–10), before the interrogative frame is read, and stays there to the end of the headline in 78–85% of cases. (ii) In a multi-question headline the lock happens inside the first sub-question 94–98% of the time; the second sub-question leaves the rank unchanged in 89–95% of headlines, and its own words encoded alone reach rank 1 only 42–58% of the time against 91–96% for the first. The word at which the first sub-question locks is the same whether it is read alone or inside the headline (95–97% identical). (iii) Numbers, dates and instrument identifiers move the embedding twice as far as ordinary content words and four times as far as interrogative words at equal position; 72–78% of all steps move toward the gold article, and the interrogative frame that closes a question displaces the vector *against* the direction the content words built in 95–99% of headlines. (iv) Clustering the rank and cosine curves yields six archetypes—instant, typical, unstable, late and never-locking paths—whose mix differs by legal area, question form and sub-question count (all $\chi^2$ $p < 10^{-8}$): real-estate and litigation headlines never lock on a number, environmental and accounting headlines do so a third of the time. (v) Read word by word, an answer retrieves its own article after 8–16 words and addresses the sub-questions in the order they were asked in 83–89% of cases. (vi) The walk is neither an independent word-embedding walk nor a wholesale rewrite at every word: the step a word contributes keeps a consistent direction across headlines (cosine 0.25–0.33 between occurrences, 0.44–0.60 for numbers, 0.00 for unrelated words), a preceding question rotates that step by about 60° and a greeting by about 30°, steps shrink as $i^{-0.8}$ under mean pooling and last-token pooling alike, and the vector of a two-question headline is reproduced to within 12–17° by a linear mix of its two questions' standalone vectors. We call this a *context-modulated additive walk*. Trajectories are presented as a token-granularity visual idiom whose ground truth is the rank in the full index rather than proximity in a projection; we show galleries of annotated trajectories for every question form and every one of the 27 legal areas (Appendix A).

## 1 INTRODUCTION

Retrieval-augmented systems treat a query as an atom: the whole string is encoded, one vector is compared with an index, the best passages are returned. Yet a query is a sequence, and a decoder-based embedder reads it left to right. Somewhere along the sequence the vector crosses from the region of the index where it is nobody's neighbour into the region where the right passage is its nearest neighbour. Where that crossing happens, whether it happens once, and which words cause it are empirical questions with practical consequences: they decide how much of a question one needs before retrieval can start, how much of it can be dropped, and whether a compound question is better sent whole or in pieces.

Vietnamese legal question answering is a good place to ask these questions. The editorial headlines of *Thư Viện Pháp Luật* (Legal

Library) [30, 38] are long (a mean of 32 words), 52% of them contain two or more questions separated by question marks, 19% open with a greeting (*Cho tôi hỏi* (Let me ask)) and 22% close with an attribution (*Câu hỏi của anh Sơn (Hà Giang)* (Question from Mr Sơn (Hà Giang))). They therefore contain, in one string, exactly the ingredients whose contribution we want to separate: a topic noun phrase, an interrogative frame, a second question, and words that carry no information at all. Decoder embedders fine-tuned for retrieval [2, 15, 26, 44] place the gold article at rank 1 for 93–96% of these headlines; the interesting question is not whether they work but *how the answer is reached.*

Text visualization has moved, over the last decade, from mapping explicit lexical statistics—tag clouds, word trees, term matrices [14]—to interpreting the latent spaces of neural language models [18]: attention routes [1, 41], layer-by-layer hidden-state geometry [3], prompt perturbation [20], and corpus cartography by non-linear projection [8]. Almost all of these idioms treat a text as a *point* (or a sequence of points across layers); the object we study is the *path* a single text traces as it is read, and its coordinate system is not a projection but the retrieval index itself: the quantity attached to every word is the rank of the correct article among 20,034, which no two-dimensional layout can distort.

This paper is about that path. Its contribution is a complete word-by-word account of one retrieval benchmark in four encoders: the embedding of every prefix of every headline (65,444 vectors per encoder), of every prefix of every sub-question read alone (61,568), and of stride-8 prefixes of 168 answers (15,839), with the rank of the gold article, its cosine, the two strongest competitors and the full vector of each prefix retained. From this we derive the *insight point* of every headline, test whether the point survives when a headline is split into its sub-questions, measure which word classes move the vector, cluster the trajectories into archetypes, map the archetypes onto the site's 27 editorial legal areas, and—because we keep every prefix vector—ask what kind of process the walk is: whether each word adds a step of its own or rewrites the vector in the light of everything before it. Every claim is illustrated with annotated trajectories drawn in the plane of the gold article and its competitors; more than seventy such panels appear in the paper, at least one for each legal area and each question form.

## 2 RELATED WORK

Dense retrievers descend from DPR [10] and Sentence-BERT [32]; the current generation fine-tunes decoder LLMs with contrastive objectives [2, 15, 22, 35] and is ranked on MTEB [5, 21] and RTEB [17], where the Nemotron-3 Embed family [26–28] and Qwen3-Embedding [44] sit near the top. Work on embedding geometry has described anisotropy and rogue dimensions [6, 7, 40], the modality gap [16] and cross-model similarity [9, 13], but treats each text as a point; the trajectory of a growing prefix has, to our knowledge, not been measured at corpus scale. Vietnamese legal retrieval has been organised around the ALQAC and Zalo challenges [4, 23, 37, 43] and studied with attentive and multi-stage models [11, 25, 29, 39]; the syllable-spaced orthography of Vietnamese [24, 42] makes "one word" a well-defined unit, which we exploit.

*Text visualization.* The taxonomy of Kucher and Kerren [14] organised the field around surface lexical features; the task-driven survey of Liu et al. [18] documents its turn toward bidirectional, model-steering analytics over dense representations. At token granularity the dominant idiom is the attribution heatmap and the attention graph [1, 36, 41]; at document granularity, narrative trajectories through a low-dimensional state space [31]; at corpus granularity, projected landscapes of sentence embeddings [8]. Hidden-state trajectories have been drawn *across layers* for a fixed input [3]; we draw them *across words* for a fixed model, and replace projected proximity by rank in the index as the primary channel. In the legal domain, visual analytics has concentrated on citation and precedent networks [12, 33], ontology-grounded norm graphs and semantic substrates that preserve institutional hierarchy [34], as surveyed by Mentzingen et al. [19]; the word-level behaviour of the neural retrievers that increasingly front such systems has not been visualised, and the present paper is a first such account.

## 3 DATA, MODELS AND THE PREFIX EXPERIMENT

### *3.1 Headlines, sub-questions and answers*

The corpus is the Q&A section of *Thư Viện Pháp Luật* (Legal Library) [30]: 20,034 articles, each a headline paired with an editorial answer that cites, quotes and glosses the governing instruments. We use the 2,144 held-out headlines of the published split as queries and the whole corpus as the index; the gold passage of a headline is its own article. A rule-based splitter cuts each headline at every question mark, strips the greeting (*Cho tôi hỏi, Xin hỏi, Nhờ anh chị giải đáp* (Let me ask, May I ask, Please advise)) and the trailing attribution, and merges fragments shorter than three words into their predecessor. This yields 3,438 sub-questions: 1,032 headlines carry one question, 946 carry two, 150 three and 16 four. We validated the splitter on 60 random headlines. Each sub-question receives one of nine interrogative forms by rule (definition, yes/no, amount/time, procedure, document, sanction, authority, what/content, other); the headline itself keeps the seven-class form label of the earlier geometry study, used here only as a covariate.

### *3.2 Encoders*

We study four decoder embedders released with open weights: Nemotron-3-Embed-8B and Nemotron-3-Embed-1B (NVIDIA, mean pooling, `query: / passage:` prefixes, 4096- and 2048-d) [26, 27], and Qwen3-Embedding-8B and Qwen3-Embedding-0.6B (Alibaba, last-token pooling, instruction prefix, 4096- and 1024-d) [44]. All run in bfloat16 with HuggingFace Transformers on one NVIDIA GB10; passages are truncated at 1,024 tokens. Colours throughout: greens for Nemotron, purples for Qwen.

### *3.3 The prefix experiment*

For every headline we form the word prefixes $w_1, w_1w_2, ..., w_1...w_n$ (capped at 64 words; 65,444 prefixes), encode each with the query prompt, and rank it against the 20,034 passage vectors. We record the rank of the gold article, its cosine, the two strongest competitors, the cosine to the previous prefix (the *step*) and to the final prefix, and keep the full fp16 vector. We do the same for every prefix of every sub-question encoded on its own (61,568 prefixes). For 168 answers stratified by form and by single/multi-question status we encode every eighth word prefix up to 800 words (15,839 prefixes) with the passage prompt and record the rank of the article's own vector, the cosine to the whole headline and to each sub-question. In total 127,012 question prefixes and 15,839 answer prefixes were encoded per encoder, 571k encodings in all.

We call the first prefix at which the gold article is rank 1 the **insight point** and report it in *content words*, i.e. after removing the greeting. A headline *holds* if it stays at rank 1 from the insight point to its last word; an *exit* is a return from rank 1 to a lower rank. Trajectories are drawn in the plane of the first two principal components of the gold vector, the twenty passages nearest the final prefix, and the path itself.

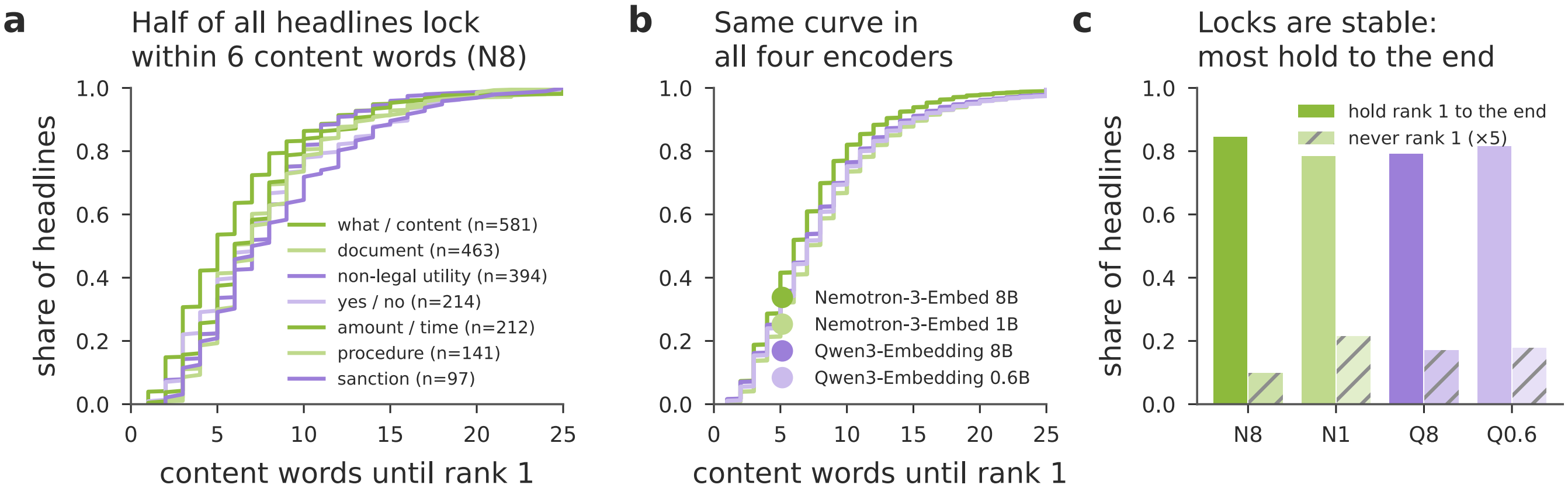

Figure 2: **The insight point.** (a) Cumulative share of headlines whose gold article is rank 1 after $k$ content words, by interrogative form (Nemotron-3-Embed-8B). (b) The same curve for the four encoders. (c) Share of headlines that hold rank 1 from the insight point to the last word, and share that never reach rank 1 (scaled ×5).

## 4 WHERE A QUESTION LOCKS ON ITS ANSWER

Figure 2 shows the distribution of the insight point. Half of all headlines have their gold article at rank 1 after six content words in Nemotron-3-Embed-8B and seven in the other three encoders; the quartiles are 4–9 words for Nemotron-3-Embed-8B, 5–11 for Nemotron-3-Embed-1B, 4–10 for Qwen3-Embedding-8B and 5–10 for Qwen3-Embedding-0.6B, and the 90th percentile lies at 13–16 words. Only 2.0% (Nemotron-3-Embed-8B), 3.4% (Qwen3-Embedding-8B), 3.5% (Qwen3-Embedding-0.6B) and 4.3% (Nemotron-3-Embed-1B) of headlines never reach rank 1 at any prefix. The forms differ less than one might expect: *what/content* headlines lock after a median of five content words in Nemotron-3-Embed-8B and six elsewhere, *amount/time* after 6–7, and the other five forms after 7–9; the interrogative frame that gives the form its name arrives after the lock and does not move it. Once locked, most headlines stay locked: 84.5% of Nemotron-3-Embed-8B paths hold rank 1 to the last word (81.7% for Qwen3-Embedding-0.6B, 79.2% for Qwen3-Embedding-8B, 78.4% for Nemotron-3-Embed-1B), with 0.16–0.22 exits per headline on average, and 92.5–96.3% of headlines finish at rank 1. Sanction headlines are the most stable (91% hold, 0.08 exits in Nemotron-3-Embed-8B), non-legal utility headlines the least (81%, 0.20).

The greeting contributes nothing. In the 406 headlines that open with *Cho tôi hỏi* (Let me ask) or a variant, the vector after the greeting alone has median rank 1,792 for the gold article and a cosine of only 0.14 to the headline's final vector; the path starts over at the first content word (grey dots at the start of the paths in Figure 14 e and Figure 15 b, Appendix A).

## 5 SPLITTING THE HEADLINE: DOES THE INSIGHT SURVIVE?

If the first six words decide the retrieval, the second question in a compound headline should be nearly inert, and reading the first question alone should reproduce the lock. Both predictions hold (Figure 3). The insight point lies inside the first sub-question for 96.6–97.9% of multi-question headlines in Nemotron-3-Embed-8B and 94–97% in the other encoders. Between the end of the first sub-question and the end of the headline the gold rank does not change for 95.1% of Nemotron-3-Embed-8B headlines (92.5% Qwen3-Embedding-0.6B, 91.0% Qwen3-Embedding-8B, 89.1% Nemotron-3-Embed-1B); the second question helps in 3.3–8.3% and hurts in 1.6–3.9%. Encoded alone, first sub-questions reach rank 1 for 95.9% (Nemotron-3-Embed-8B), 94.1% (Qwen3-Embedding-0.6B), 93.3% (Qwen3-Embedding-8B) and 90.5% (Nemotron-3-Embed-1B) of headlines, second sub-questions for 57.9%, 41.7%, 46.1% and 50.7%, third for 25–40%. Among two-question headlines in Nemotron-3-Embed-8B, both halves retrieve the article alone in 54.9%, only the first in 41.5%, only the second in 1.8% and neither in 1.8%; the Qwen models tilt further toward "first only" (50.6% and 55.1%). A failing second sub-question is not far off—its median rank is 4 (Nemotron-3-Embed-8B), 5 (Nemotron-3-Embed-1B), 7 (Qwen3-Embedding-8B) and 15 (Qwen3-Embedding-0.6B) and 67–83% of them are within the top ten—but it is the first question that names the topic and the second that presupposes it. The failure rate of second sub-questions is flat across their forms (50–66% rank 1 in Nemotron-3-Embed-8B), so it is position, not form, that matters.

Context does not move the insight word. For the 2,084 first sub-questions with a lock in both readings, the content word at which the sub-question locks alone is identical to the one at which the full headline locks in 95.6% of cases (Nemotron-3-Embed-8B; 96.6% Nemotron-3-Embed-1B, 95.2% Qwen3-Embedding-8B, 94.5% Qwen3-Embedding-0.6B) and within one word in 98–99%. The prefix vector is a function of the prefix, not of what follows, and the greeting that precedes the first content word is, as shown above, forgotten by the second one. Figure 4 shows nine compound headlines with the isolated sub-question paths overlaid: the first sub-question's path (thin green) retraces the headline's path; the second sub-question's path (thin blue) starts from a different corner of the plane and, more often than not, ends in the competitor cloud.

## 6 WHAT MOVES THE VECTOR

Every word shifts the vector, but not equally (Figure 5). Controlling for position (words 5 onward), a word containing a digit—a date, an amount, an instrument number such as *29/2024/TT-BCT*—moves the Nemotron-3-Embed-8B vector by 0.465 on the unit sphere, an ordinary content word by 0.245, an interrogative or function word (*như thế nào, gì, có, không* (how, what, yes, no)) by 0.115 and an attribution word by 0.113; the ratios are the same in the other encoders (0.394/0.219/0.109/0.096 for Qwen3-Embedding-8B). The words with the largest mean steps in the corpus are dates (*1/7/2026*: 0.77; *01/01/2026*: 0.62) and price-of-petrol vocabulary (*Petrolimex*, *xăng*); the smallest are *mong*, *thế*, *như* and *câu* (0.06–0.08), the words of the frame and the attribution. Steps shrink along the headline for all classes (Figure 5 b): the first word moves the vector by 0.9, the tenth by 0.35, the thirtieth by 0.15, so the space becomes progressively harder to leave.

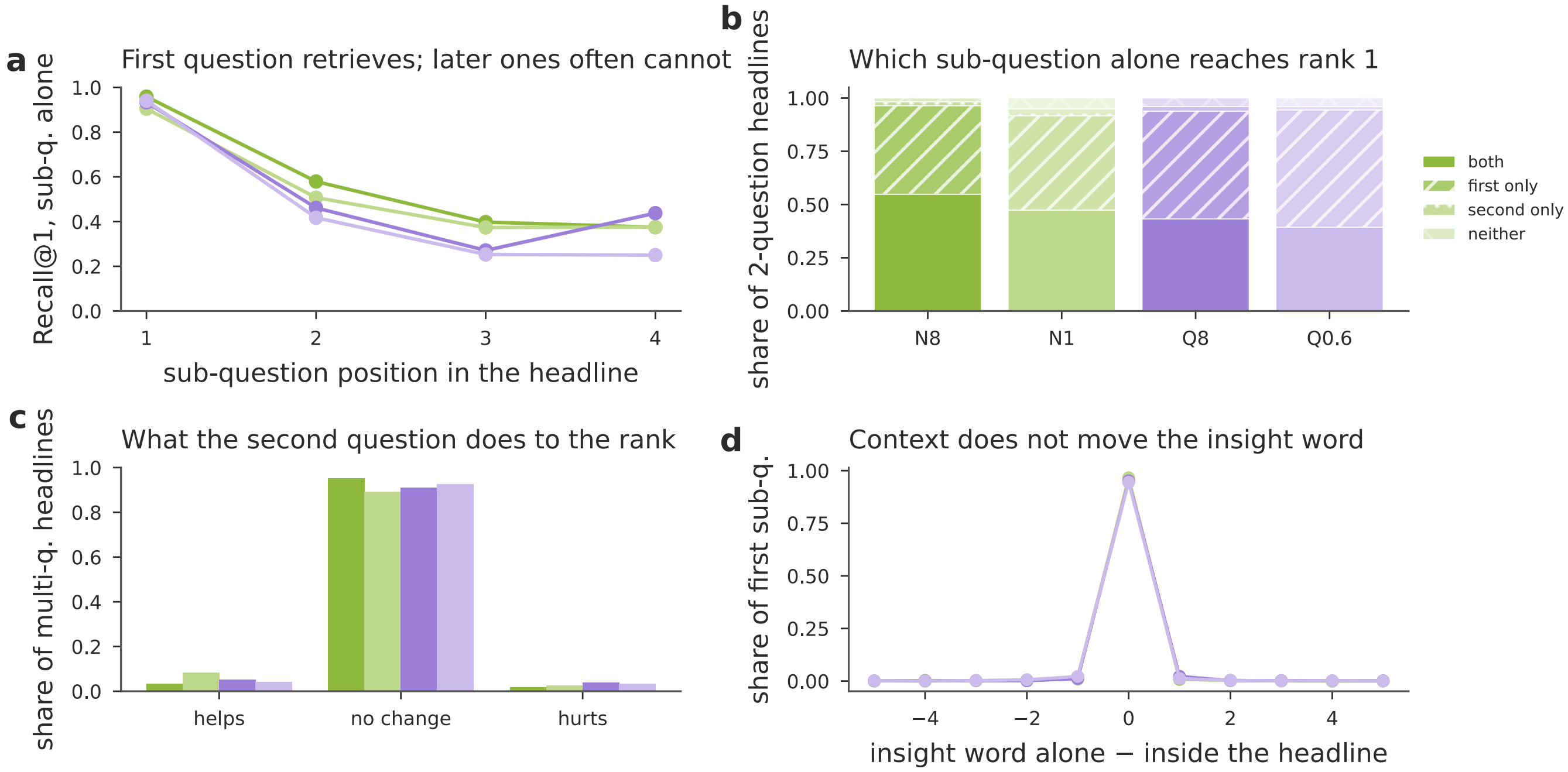

Figure 3: **Sub-questions alone and inside the headline.** (a) Recall@1 of each sub-question encoded alone, by its position in the headline. (b) For two-question headlines: whether the first, the second, both or neither sub-question alone reaches rank 1. (c) Change of the gold rank between the end of the first sub-question and the end of the headline. (d) Difference between the insight word of the first sub-question read alone and inside the headline.

Three vector-level facts hold in every encoder. First, 72–78% of all steps have a positive component along the direction to the gold article (77.9% Nemotron-3-Embed-8B, 73.5% Nemotron-3-Embed-1B, 74.8% Qwen3-Embedding-8B, 72.1% Qwen3-Embedding-0.6B): the path is a noisy but directed walk. Second, the paths are tortuous—their length is 5.3–6.0 times the net displacement from first to last prefix—but Nemotron-3-Embed-8B turns least, with 1.3 direction reversals per headline against 3.0 (Qwen3-Embedding-0.6B), 3.6 (Nemotron-3-Embed-1B) and 4.5 (Qwen3-Embedding-8B). Third, the interrogative frame makes a U-turn: the displacement accumulated after the insight point up to the end of the first sub-question has a negative cosine with the displacement from the first content word to the insight point in 98.7% of Nemotron-3-Embed-8B headlines (mean −0.23; 95.2% for Nemotron-3-Embed-1B, 97.6% for Qwen3-Embedding-8B, 97.1% for Qwen3-Embedding-0.6B, means −0.19 to −0.20). The frame words pull the vector partly back toward where the question started, which is why grey words in the galleries so often double back along the green path.

## 7 ADDITIVE OR AUTOREGRESSIVE? WHAT KIND OF WALK THIS IS

The galleries invite a question that the rank curves cannot answer: is the path an *independent walk*, in which every word adds a step of its own that is the same wherever the word occurs, or an *autoregressive* process, in which each new word rewrites the vector in the light of everything before it? The two encoder families make the question sharp. Both are causal decoders, so a token's hidden state depends on the tokens before it and on nothing after. Nemotron pools by *mean*, so before normalisation the prefix vector is literally a running sum of the token states and the step of word $i$ is that word's own contextual state divided by $i$: additive by construction, with all the context dependence hidden inside the token state. Qwen pools by *last token*, so the whole vector is rewritten at every word and nothing forces adjacent prefixes to be related at all. Four tests on the stored prefix vectors (Figure 6) give the same answer for both families.

*A word's step keeps part of its direction, not all of it (Figure 6 a).* For the 399 words that occur at least thirty times at position five or later (43,937 occurrences in Nemotron-3-Embed-8B), the cosine between the steps of the same word in two different headlines averages 0.25 (Nemotron-3-Embed-8B), 0.32 (Nemotron-3-Embed-1B), 0.27 (Qwen3-Embedding-8B) and 0.33 (Qwen3-Embedding-0.6B); between steps of different words it is 0.00. So a word does carry a direction of its own—an angle of 71–76° between two occurrences, against 90° for chance—but that direction accounts for less than a tenth of the step's variance. The share is a property of the word class: interrogative and function words keep almost nothing (0.13–0.20; *là, và, nào, của* (is, and, which, of) are at 0.04–0.05), content words keep a quarter to a third, and numbers, dates and instrument identifiers keep half to two thirds (0.44–0.60; the date *1/7/2025* reaches 0.75). The word ranking is shared across families (Spearman 0.87 between Nemotron-3-Embed-8B and Qwen3-Embedding-8B): the same words are context-free in every encoder.

*A preceding question rotates every step by about 60°; a greeting by about 30° (Figure 6 b, Figure 7).* The second sub-question of a compound headline is encoded twice: alone, and inside the headline after the first question. The words are identical; only the left context differs. Step by step, the two paths agree with a mean cosine of 0.50 (Nemotron-3-Embed-8B), 0.62 (Nemotron-3-Embed-1B),

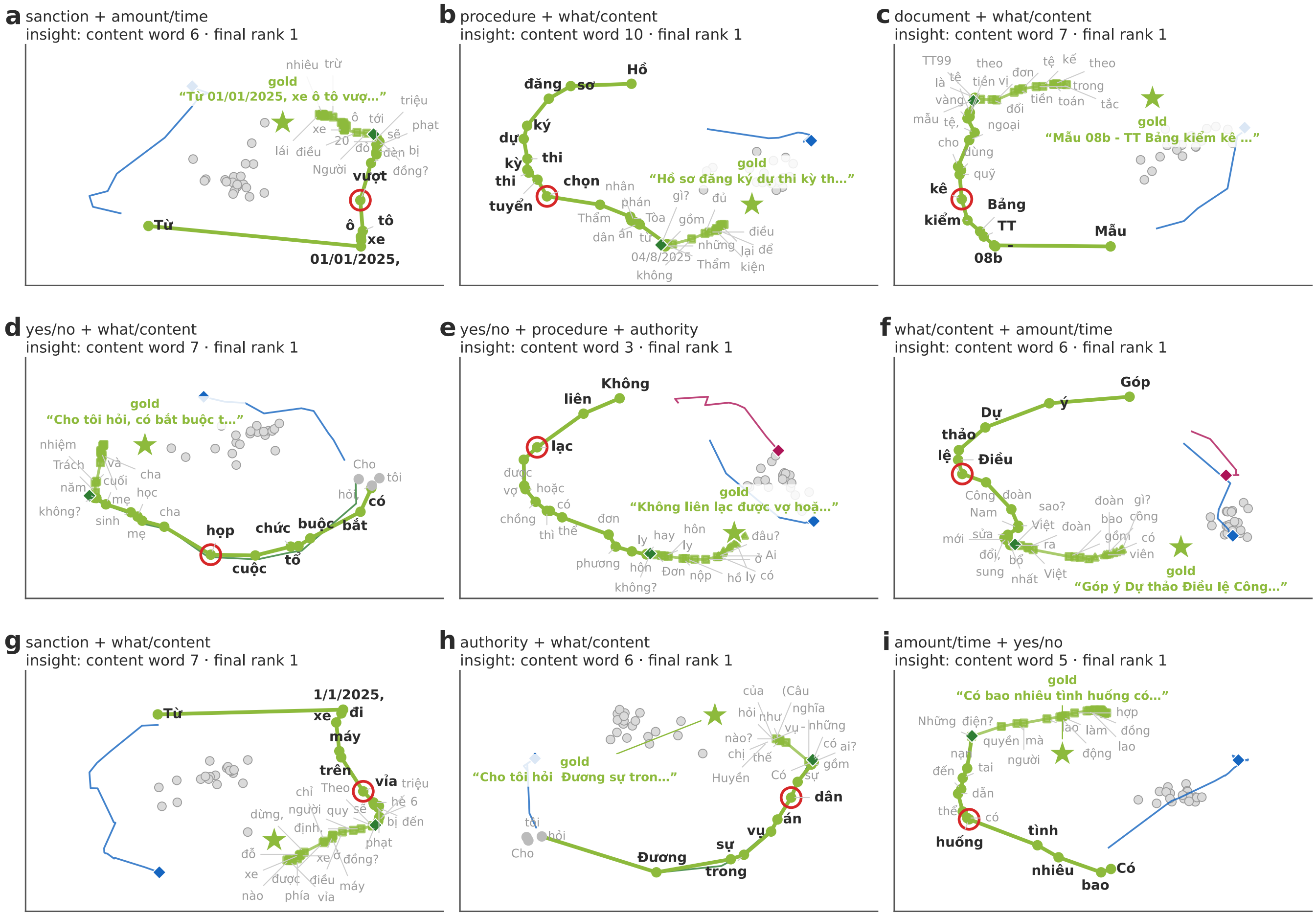

Figure 4: **Nine multi-question headlines with their sub-questions read alone** (Nemotron-3-Embed-8B). Circles: first sub-question inside the headline; squares: second sub-question inside the headline; thin green and blue lines: the first and second sub-questions encoded on their own, ending at a diamond. The second question alone lands away from the gold article in most panels.

0.45 (Qwen3-Embedding-8B) and 0.54 (Qwen3-Embedding-0.6B)—a rotation of 52–64° per step—and the in-context steps are a third as long (median ratio 0.33–0.38) because they arrive later in the sequence. The control is the first sub-question, whose only extra context inside the headline is a greeting: there the cosine is 0.85–0.92 (24–31°), and when there is no greeting the two texts are identical and the cosine is 1.00 in Nemotron and 0.97–0.99 in Qwen, the latter being the bfloat16 batch-composition noise floor of last-token pooling. Rotation grows with the length of the context that precedes: 0.59 when the first question is at most six words, 0.47 when it exceeds fifteen (Nemotron-3-Embed-8B). It is smallest for yes/no second questions (0.54) and largest for definitions (0.43), whose *là gì* frame is the most context-bound. The shape of the whole path changes accordingly: after translation and scaling, the distance between the alone and in-context paths is 0.90–1.03 for second questions against 0.40–0.50 for the greeting control (0 identical, 1.41 unrelated). Figure 7 shows three such pairs.

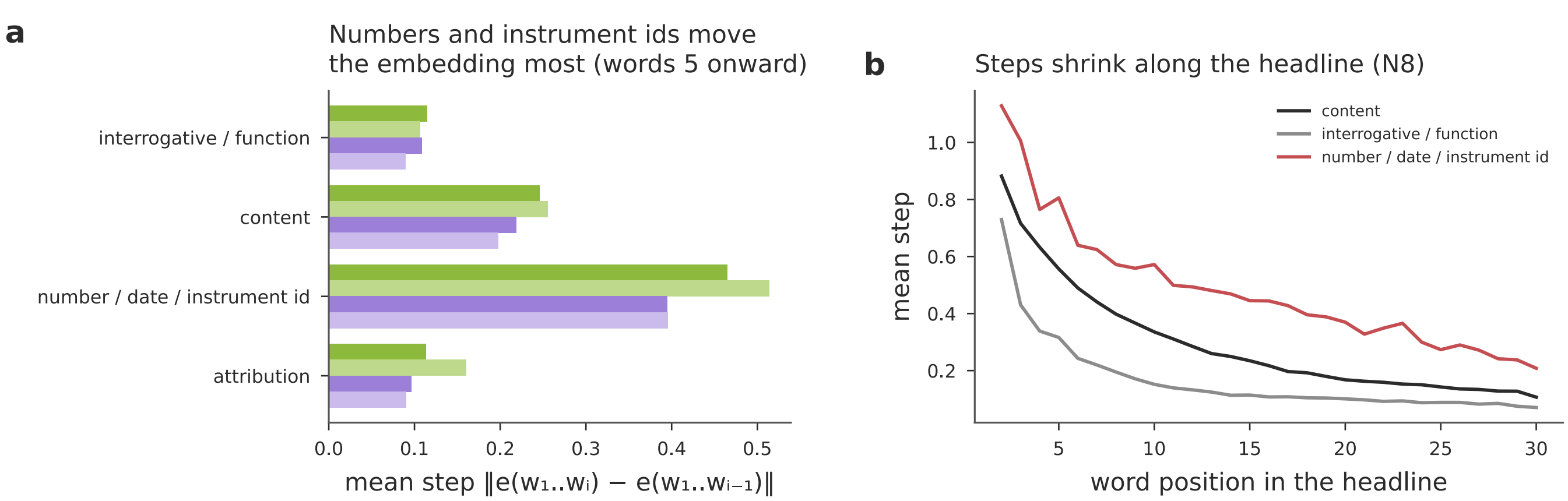

Figure 5: **Word-level steps.** (a) Mean step length $\|e(w_1..w_i) - e(w_1..w_{i-1})\|$ by class of the word $w_i$, for words at position 5 or later so that the early-position effect is removed. (b) Mean step by position in the headline for three word classes (Nemotron-3-Embed-8B).

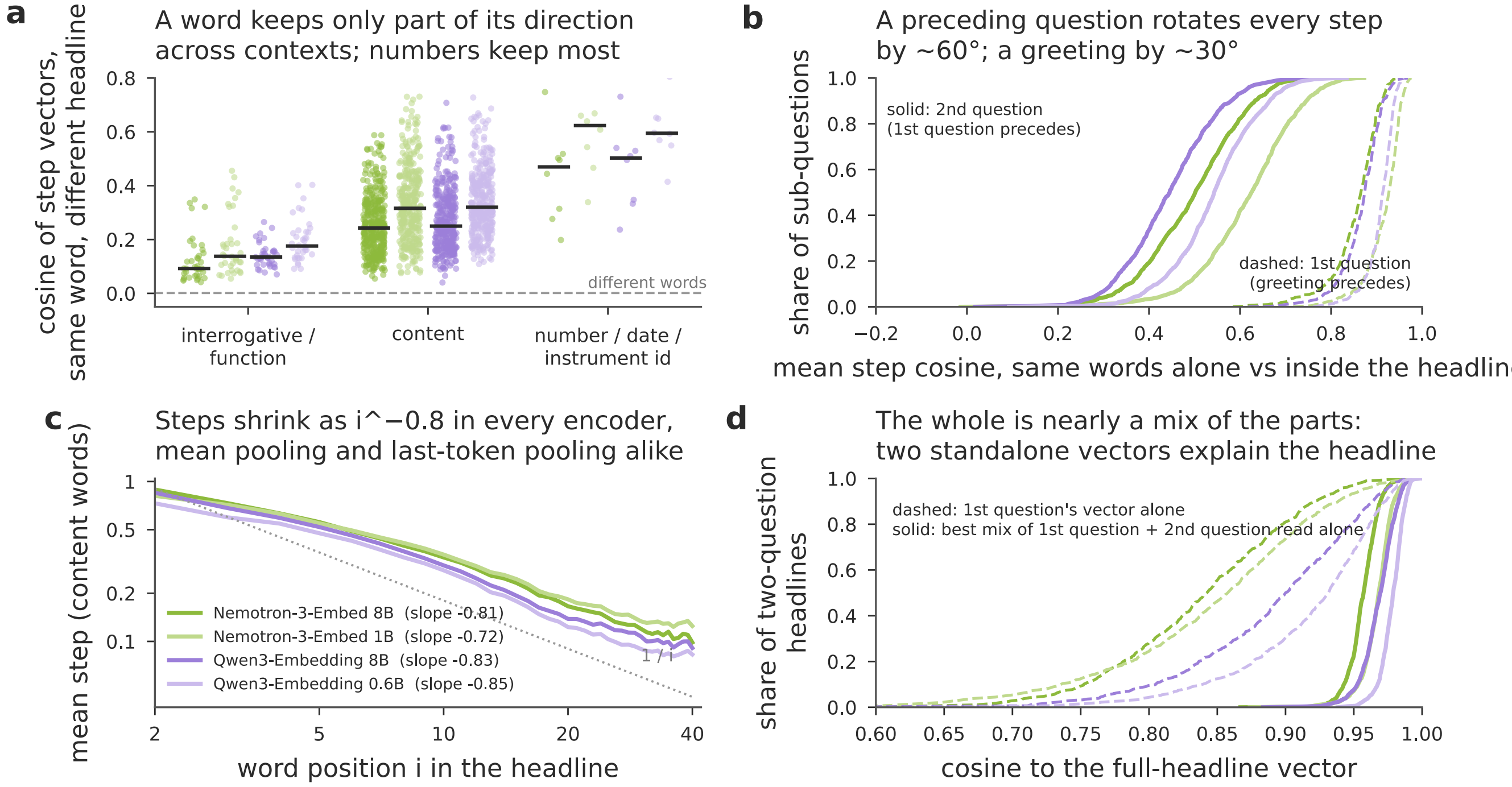

Figure 6: **Four tests of what kind of process the walk is.** (a) Cosine between the step vectors $e(w_1..w_i) - e(w_1..w_{i-1})$ of the same word $w_i$ in different headlines (399 words with ≥30 occurrences at position ≥5; one dot per word, bars: medians); the dashed line is the cosine between steps of different words. (b) For every sub-question, the mean cosine between its steps read alone and the same steps inside the headline: solid, second sub-questions (a first question precedes; $n$=1,262); dashed, first sub-questions preceded only by a greeting ($n$=406). (c) Mean step length of content words against word position, log–log, with the fitted slope and a $1/i$ reference. (d) For 922 two-question headlines, cosine between the full-headline vector and the first question's vector (dashed) or the best linear mix of the first question's vector and the second question's standalone vector (solid).

*Steps shrink as a power law whatever the pooling (Figure 6 c).* Mean pooling predicts that the step of word $i$ scales as $1/i$; last-token pooling predicts nothing. Measured on content words, the mean step falls from 0.73–0.89 at the second word to 0.28–0.35 at the tenth and 0.09–0.13 at the thirtieth, with log–log slopes of −0.81 (Nemotron-3-Embed-8B), −0.72 (Nemotron-3-Embed-1B), −0.83 (Qwen3-Embedding-8B) and −0.85 (Qwen3-Embedding-0.6B). The two Qwen models, which have no averaging in their read-out, shrink at least as fast as the two Nemotron models that do. The shrinkage is therefore not an artefact of pooling; it is a property of the contextual token states themselves, whose sensitivity to one more word decays as the context lengthens. The exponent being shallower than −1 says that later words are, per token, slightly *more* influential than a plain average would make them.

*The whole is nearly a mix of the parts (Figure 6 d).* For the 922 two-question headlines whose full text fits the 64-word cap, we ask whether the headline vector can be written as $\alpha \cdot e(\text{question}_1) + \beta \cdot e(\text{question}_2 \text{ alone})$. The first question's vector alone has cosine 0.84–0.92 with the headline (24–33°); the best two-term mix reaches 0.96–0.98 (12–17°), with coefficients of about 0.65 on the first question and 0.40–0.50 on the second in every encoder. A vector for a compound question can thus be assembled, to within a small angle, from the vectors of its questions encoded separately—even in the Qwen models, whose read-out contains no sum.

Taken together: the walk is a *context-modulated additive walk*. Each word contributes a step whose direction is anchored to the word (strongly for numbers and topical nouns, weakly for the frame), rotated by the words that precede it, and shortened by their number; consecutive steps are uncorrelated in Nemotron-3-Embed-8B (mean cosine 0.00) and mildly anti-correlated in the Qwen models (−0.05 to −0.09; 63–68% of turns exceed 90°), which is the zigzag that makes their paths more tortuous. The result is the same in the family whose architecture makes additivity exact and in the family whose architecture makes no such promise, so the additivity is learned, not built in.

## 8 SIX ARCHETYPES, AND WHERE THEY LIVE

Clustering the resampled rank, cosine and step curves of the 2,144 Nemotron-3-Embed-8B trajectories gives six archetypes (Figure 8). **Instant lock** (636 headlines, 30%): the first content word already places the gold article near rank 300 and the lock comes at content word 4; 91% of these are multi-question headlines whose first word is the topic noun. Two **typical** clusters (649 and 427 headlines) lock at 7–8 content words from a start near rank 3,000–5,000 and hold in 87–88% of cases; the smaller one is almost entirely single-question (94%) and greeting-rich (29%). **Unstable lock** (288, 13%): the lock comes at word 6 but holds only 67% of the time, with 0.35 exits per headline and 12% of paths finishing below rank 1—the paths that wander back into the competitor cloud when the frame or a second question is read. **Late lock** (122, 6%): the first word ranks the gold near 8,600 and rank 1 arrives only after 13 content words, two-thirds of the way through the content span; these are single questions whose topic is generic until a qualifier arrives. **Never** (22, 1%): 82% never reach rank 1.

The archetype mix is not uniform. It depends on the number of sub-questions ($\chi^2$ $p \approx 10^{-204}$), on the question form ($p = 8\times10^{-9}$) and on the legal area ($p = 1.3\times10^{-10}$). Transport (*Giao thông – Vận*

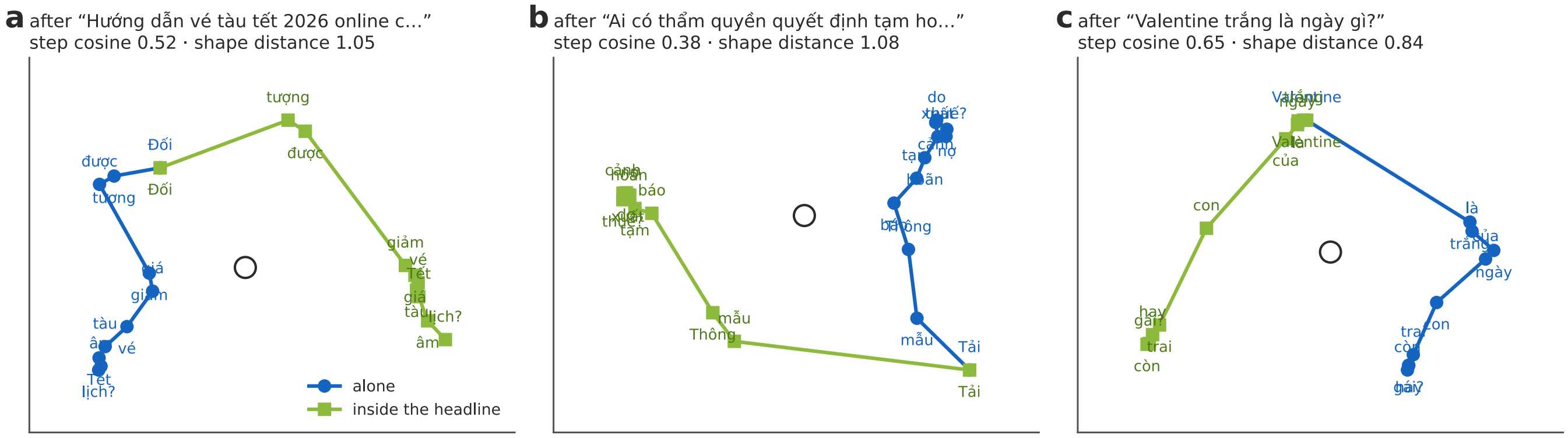

Figure 7: **The same words, two contexts** (Nemotron-3-Embed-8B). Three second sub-questions read alone (blue) and inside their headline after the first question (green), drawn from a common origin in the plane of the two paths after scaling each to unit length; every word is written at its own step. The title gives the first question that precedes and the two statistics of Figure 6 b. The words are identical; the shape is not.

*tải* (transport), 47% instant), culture and society (44%) and public administration (40%) are dominated by instant locks; securities (12%), import–export (15%) and litigation procedure (15%) are not. Figure 9 maps four statistics over the 27 areas. The median insight point varies only between 5 and 8 content words, and the hold rate between 75% (*Xây dựng – Đô thị* (construction and urban planning)) and 95% (*Tài chính nhà nước* (public finance)), but the share of headlines whose insight word is a number differs by an order of magnitude: 0% in real estate and litigation procedure, 2–3% in intellectual property, administrative violations and legal services, against 21% in criminal liability, 29% in accounting and audit and 32% in natural resources and environment (*Tài nguyên – Môi trường* (natural resources and environment)). In the number-locking areas the discriminating information is an instrument or a date (*Thông tư 29/2024/TT-BCT, Nghị định 70* (Circular 29/2024, Decree 70)); in real estate it is the noun phrase. Number locks are as stable as word locks (85% hold in both groups) but arrive one word earlier (median 5 against 6).

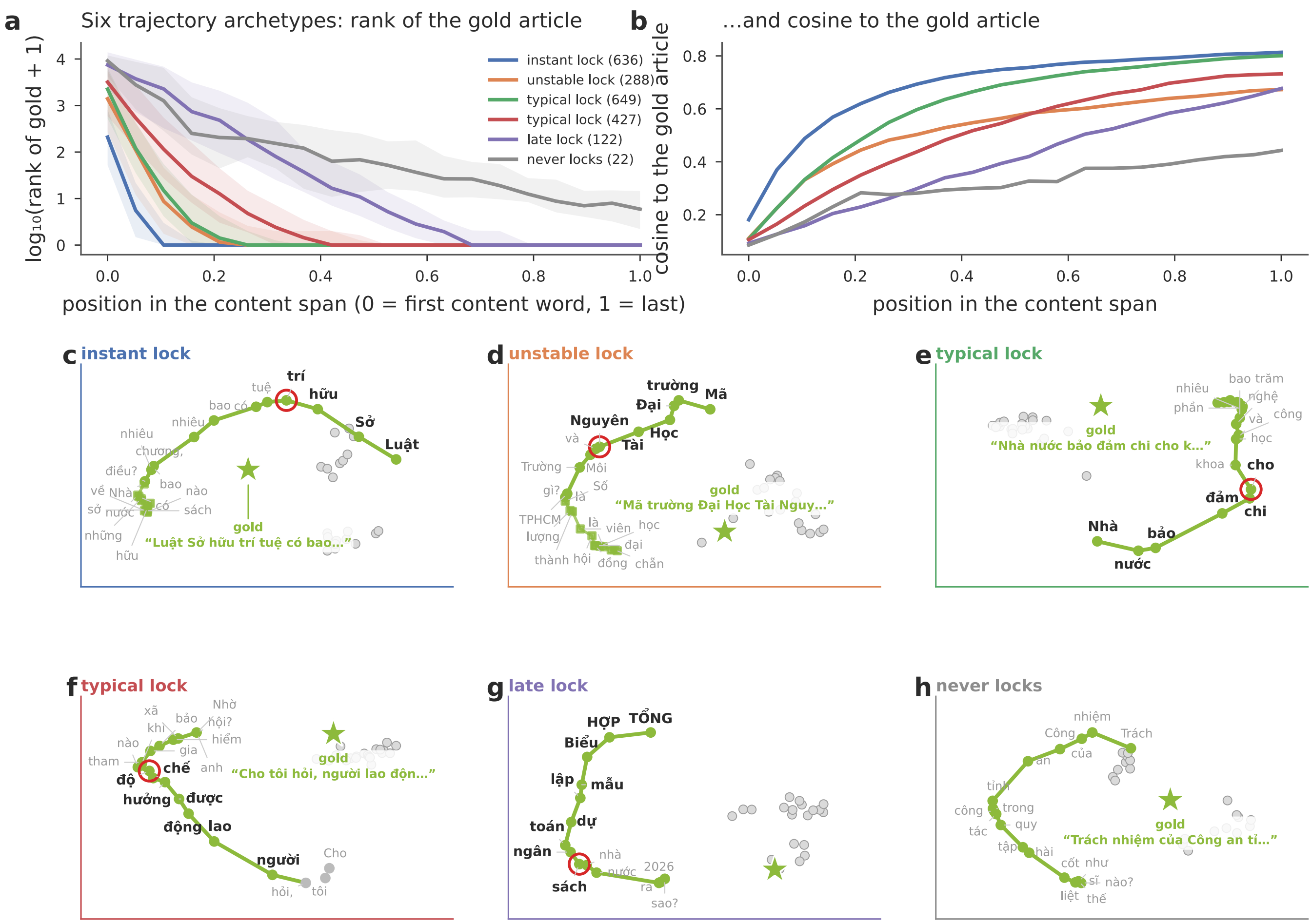

Figure 8: **Trajectory archetypes** (Nemotron-3-Embed-8B). Each headline's rank curve, cosine-to-gold curve and step curve over its content span were resampled to 20 points and clustered (k-means, k = 6). (a) Median log-rank curve of each cluster with interquartile band; (b) median cosine to the gold article. (c–h) The medoid headline of each cluster, word-annotated.

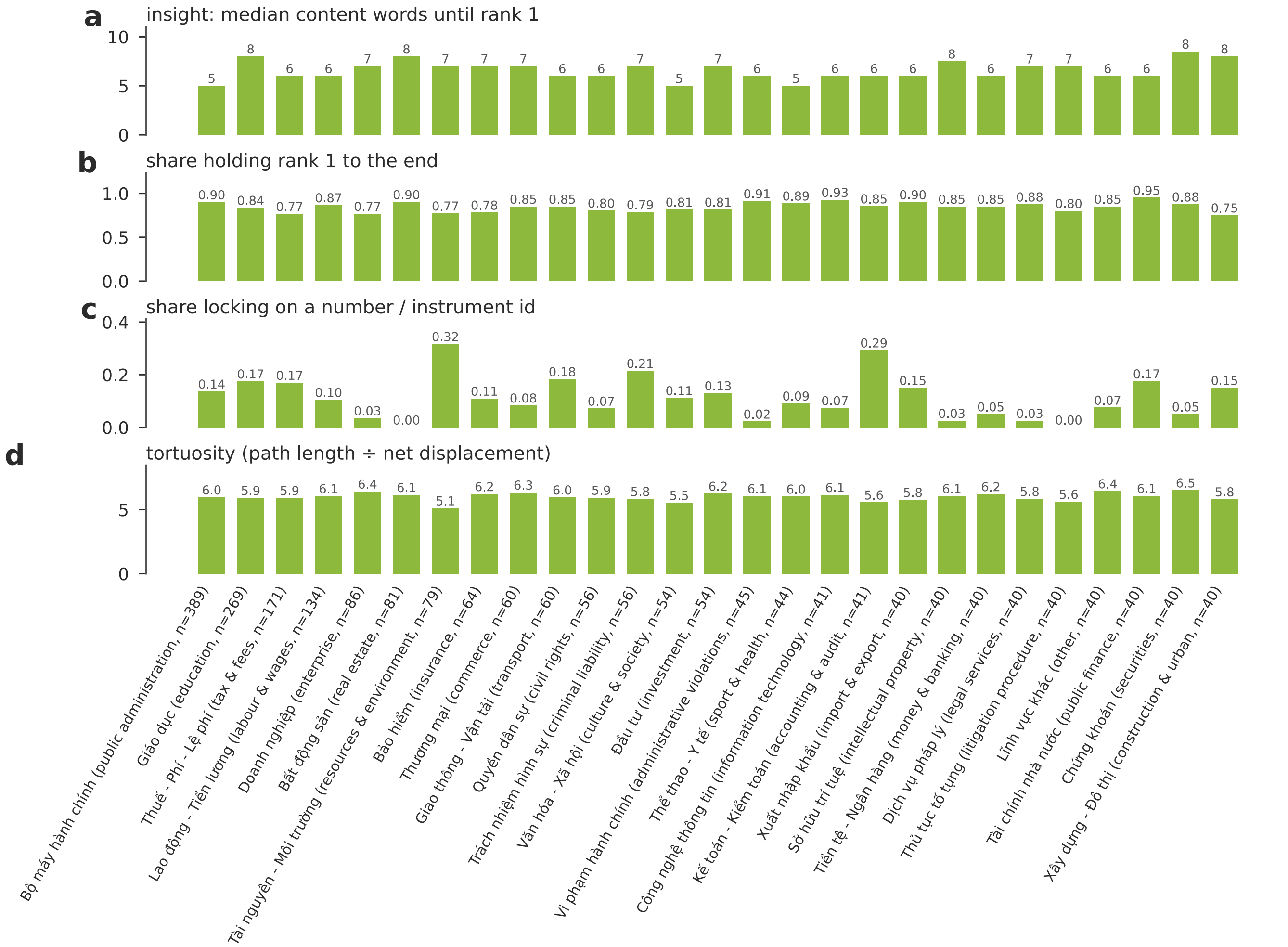

Figure 9: **Trajectory statistics by editorial legal area** (Nemotron-3-Embed-8B, 27 areas ordered by size). (a) Median insight point in content words; (b) share of headlines holding rank 1 to the end; (c) share whose insight word contains a digit (a date, amount or instrument number); (d) median tortuosity.

## 9 THE ANSWER READ WORD BY WORD

Read the same way, an answer behaves like its question (Figure 10). Used as a query, its own first 8 words (median; 16 for Nemotron-3-Embed-1B) already retrieve the article at rank 1—the restated heading is the whole signal. Its similarity to the headline peaks where the form predicts: at 14–27% of the encoded text for document answers, whose first lines name the template, and at 63–75% for yes/no and 51–71% for sanction answers, whose conclusion (*Như vậy* (Thus)) states the verdict or the fine. Compound answers address the questions in the order they were asked: the prefix most similar to the first sub-question precedes the one most similar to the second in 89.3% of multi-question answers for both Nemotron models, 85.7% for Qwen3-Embedding-8B and 83.3% for Qwen3-Embedding-0.6B, with a median gap of 35–54% of the answer. The encoders differ in how they converge: the Qwen prefix reaches 90% cosine to the full-answer vector after 112 (Qwen3-Embedding-8B) and 48 (Qwen3-Embedding-0.6B) words, the Nemotron prefix after 232 and 304, because last-token pooling settles once the topic is stated while mean pooling keeps averaging in every new word; the Qwen answer paths are correspondingly rougher (tortuosity 10.1 for Qwen3-Embedding-8B against 5.5–6.6).

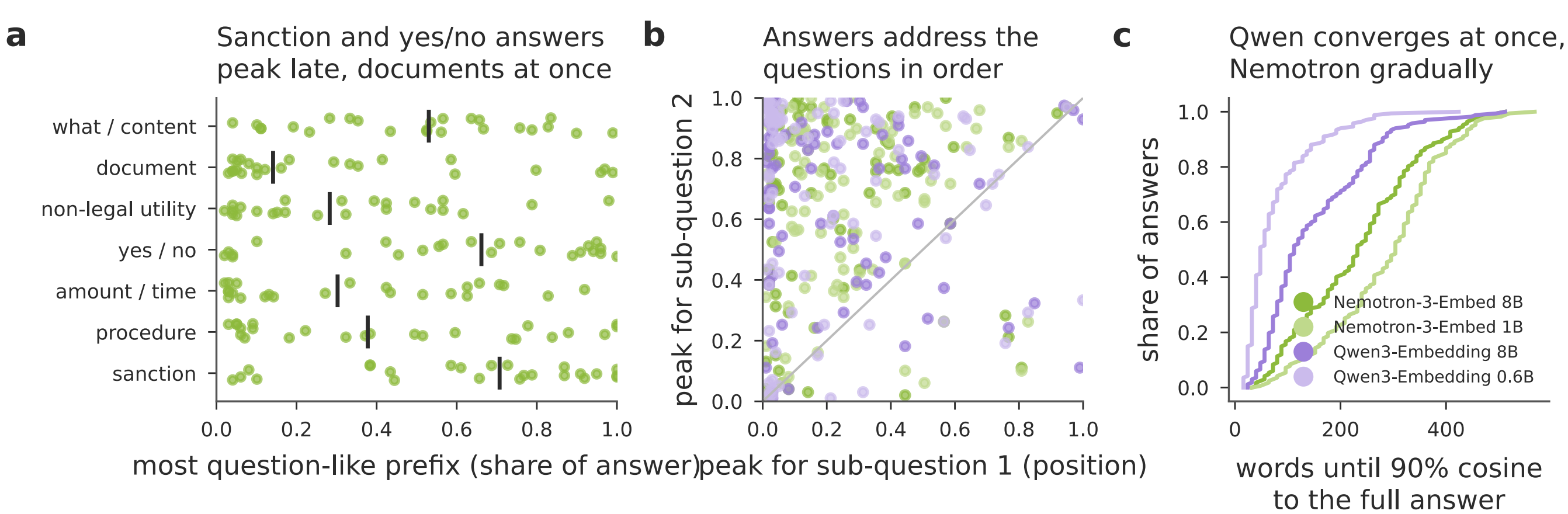

Figure 10: **Answer walks** (168 answers, stride 8 words, up to 800 words). (a) Position, as a share of the encoded answer, of the prefix most similar to the headline, by form (Nemotron-3-Embed-8B; bars: medians). (b) For multi-question answers, the position of peak similarity to the first sub-question against the second; points above the diagonal address the questions in order. (c) Words read until the prefix reaches 90% cosine to the full answer vector.

## 10 DO THE ENCODERS AGREE?

The distributions agree; the individual insight words agree only partly. Where both encoders find a lock, Nemotron-3-Embed-8B and Qwen3-Embedding-8B lock on the same word for 42.9% of headlines and within one word for 65.6%; Nemotron-3-Embed-8B and Nemotron-3-Embed-1B for 39.7% and 63.7%; Qwen3-Embedding-0.6B and Qwen3-Embedding-8B for 42.5% and 66.5%; the cross-family small-model pairs for 34% and 58%. All four lock on the same word in 16% of headlines and within one word in 37%. The insight point is thus a property of the headline to within a word or two, not to the word; the differences are in how many qualifiers an encoder needs before the topic noun phrase separates the article from its siblings, and Nemotron-3-Embed-8B needs the fewest.

## 11 IMPLICATIONS

**Retrieval can start early.** Half of all headlines are resolved after six content words and 90% after 13–16; a streaming or speculative retriever can issue its first index probe as soon as the topic noun phrase is complete and treat the interrogative frame as confirmation, not information. **Truncate from the end, and drop the greeting.** The greeting is forgotten by the second content word; the frame moves the vector against the direction it came from; the attribution moves it hardly at all. A query budget should spend its tokens on the first sub-question. **Split compound questions for the second question, not the first.** The first sub-question alone reproduces the headline's retrieval in 91–96% of cases; the second alone reaches rank 1 in fewer than 60% because it presupposes the first. A RAG system that answers a compound question should retrieve for the first sub-question and re-query for the second with the first's topic prepended. **Numbers are heavy.** A date or instrument number moves the vector twice as far as any other word and, in a third of environmental and accounting headlines, is the word that decides retrieval; systems that normalise or strip numerals before embedding remove the lock. **Answers are ordered.** Because an answer addresses sub-questions in order, chunking by sub-question boundary is sound, and the chunk that answers the first question is usually in the first half. **Prefix vectors compose.** Because the walk is additive to within a small angle, the vector of a compound query can be approximated from cached vectors of its parts (0.96–0.98 cosine), and because both families are causal decoders, extending a query by one word reuses the key–value cache of the prefix: a streaming retriever that re-probes the index after every word pays one token of compute per word, not one query. Because the steps are context-modulated, however, a cache keyed on words rather than prefixes would not work—the step of *mẫu* (template) after one question is not its step after another.

## 12 LIMITATIONS

Our gold label is the headline→article pair of one publisher, so rank measures self-retrieval of an editorial headline, not general question answering. Word boundaries are syllable spaces, so a Vietnamese compound is read as two or three steps. The splitter and the sub-question forms are rule-based; the interrogative forms were validated on 60 headlines and are covariates, not labels. Answer walks use a stride of eight words and 168 answers. The 2-D planes are per-panel PCA projections chosen to show the gold, its competitors and the path; distances between panels are not comparable, and a star that appears far from the path's end is a projection artefact when the final rank is 1. Prefixes were encoded with the query prompt; a passage prompt would trace a different path. The additivity tests compare vectors of prefixes and of sub-questions encoded separately; they measure how far the whole departs from a linear mix of its parts, not the mechanism inside the network that produces the departure.

## 13 CONCLUSION

Reading 2,144 legal headlines one word at a time through four decoder embedders shows a single behaviour with a few variants: the vector walks, mostly toward the answer, and locks on it after the topic noun phrase—six or seven content words—before the question has been asked. The frame turns the vector back a little but does not unlock it; a second question is read but rarely matters; a first question read alone locks on the same word it locks on inside the headline. Numbers and instrument identifiers are the heaviest words and, in some areas of law, the deciding ones. The answers, read the same way, retrieve themselves after a dozen words and address the questions in order. The walk itself is a context-modulated additive process: a word's step keeps a direction of its own, is rotated by about 60° by a preceding question, shrinks as $i^{-0.8}$ whatever the pooling, and sums, to within 12–17°, to the vector of the whole—in the family whose read-out is a sum and in the family whose read-out is not. The full prefix vectors, ranks and tables for all 127,012 question prefixes and 15,839 answer prefixes in the four encoders accompany the paper.

## A GALLERIES

These galleries show trajectories for every question form (six headlines each, three single- and three multi-question where available) and every legal area, drawn from the full vectors of all 2,144 headlines; no headline appears twice in the paper. In each panel the gold article is the star, the twenty passages nearest the final prefix are grey, the headline path is green with one dot per word (squares mark a second sub-question), the red ring is the insight point, and thin blue lines are second sub-questions encoded alone. Titles give the forms of the sub-questions, the insight point in content words and the final rank.

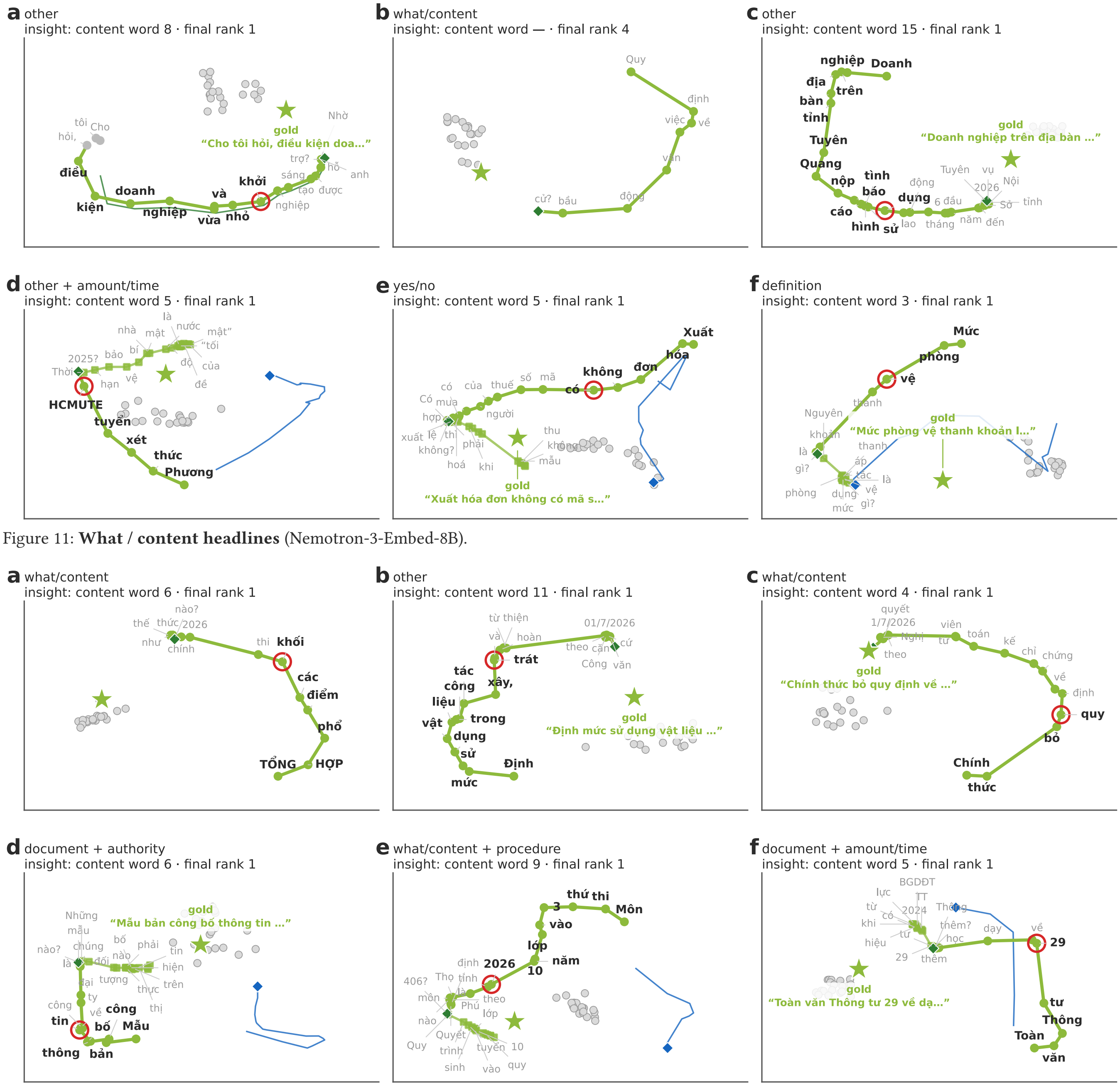

Figure 11: **What / content headlines** (Nemotron-3-Embed-8B).

Figure 12: **Document-request headlines** (Nemotron-3-Embed-8B): the template name locks the article; *mẫu, tải* (template, download) add nothing.

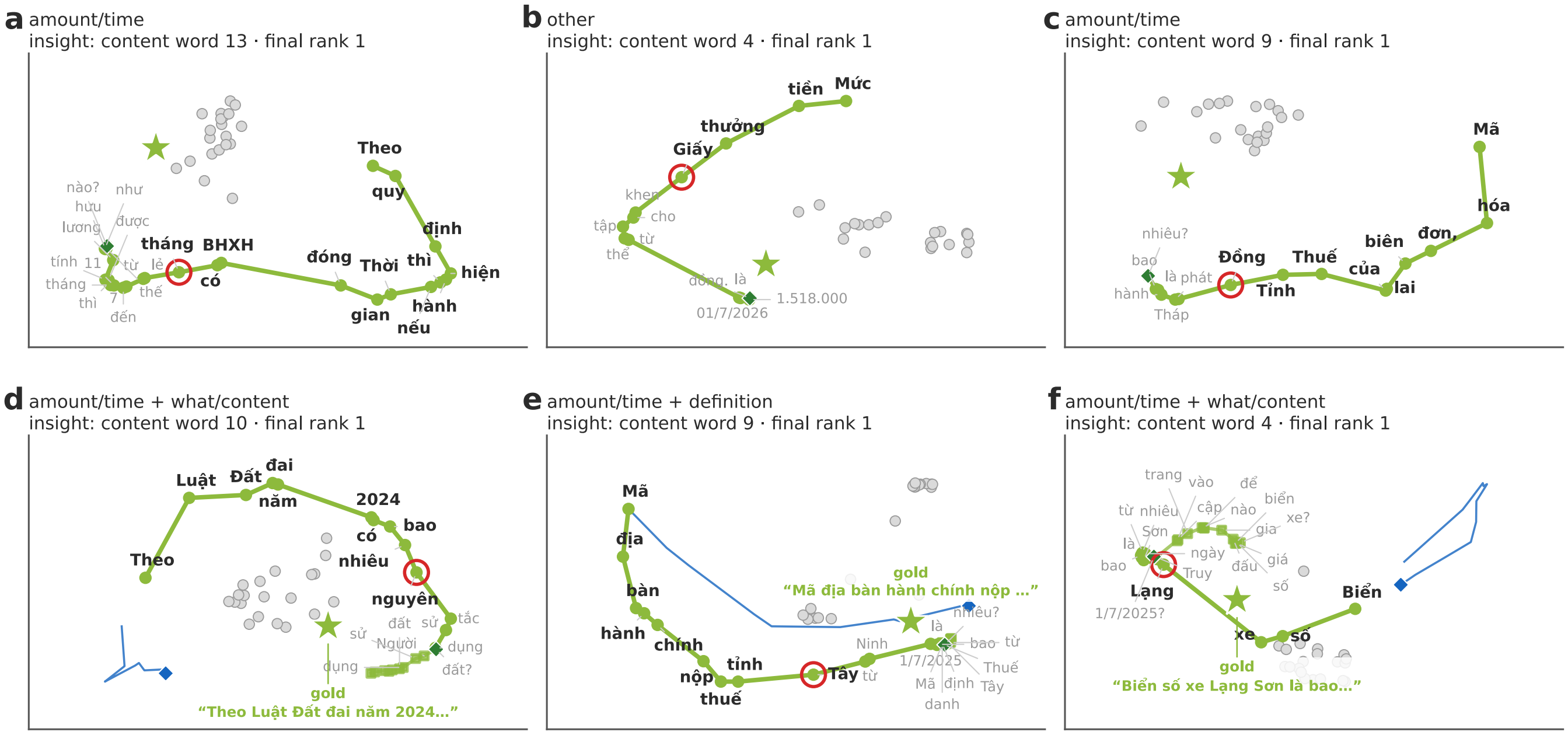

Figure 13: **Amount / time headlines** (Nemotron-3-Embed-8B): the quantity asked for is in the frame, the lock is on the subject.

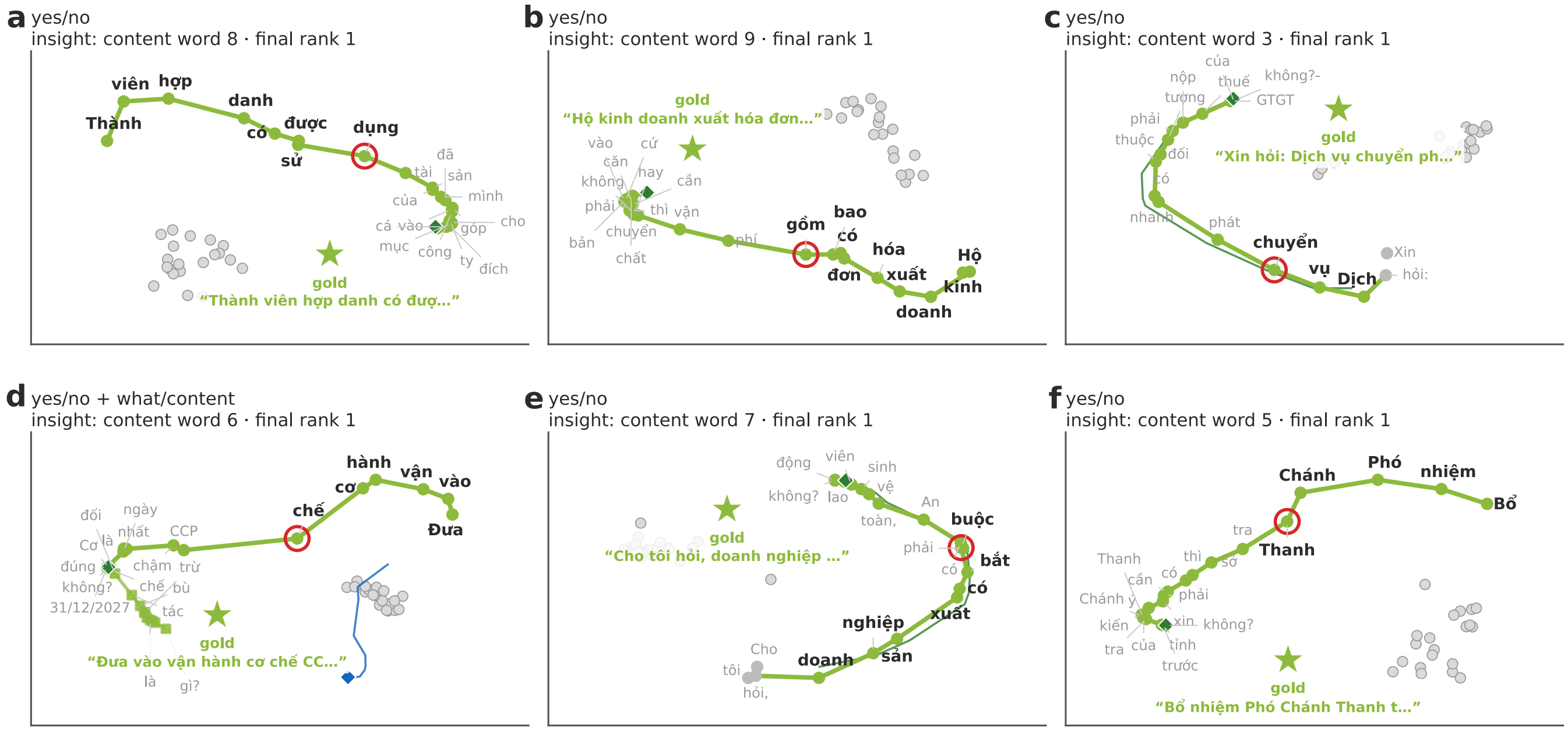

Figure 14: **Yes / no headlines** (Nemotron-3-Embed-8B): (e) opens with a greeting (grey) that the path leaves at once.

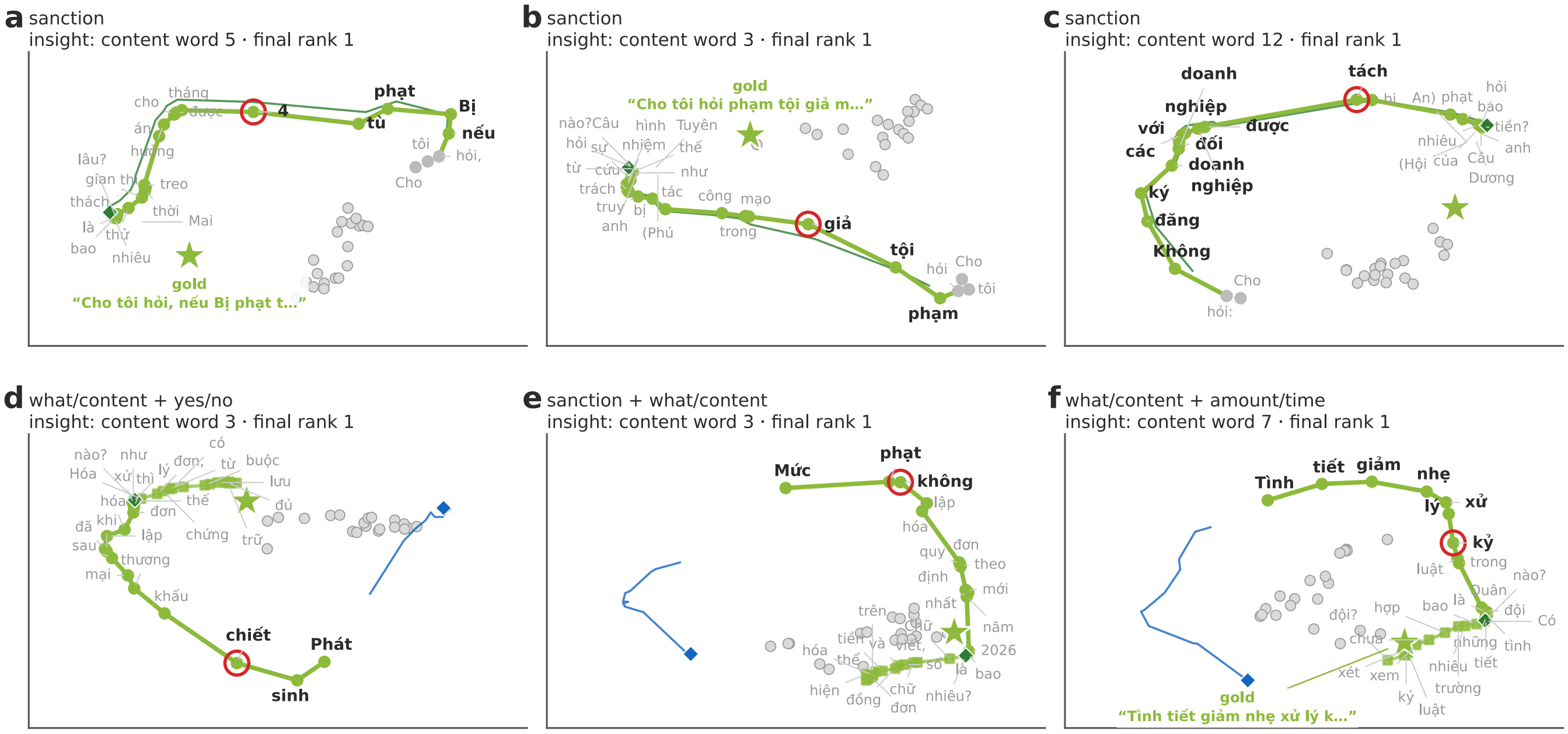

Figure 15: **Sanction headlines** (Nemotron-3-Embed-8B): the most stable form; (b) opens with *Cho tôi hỏi* (Let me ask).

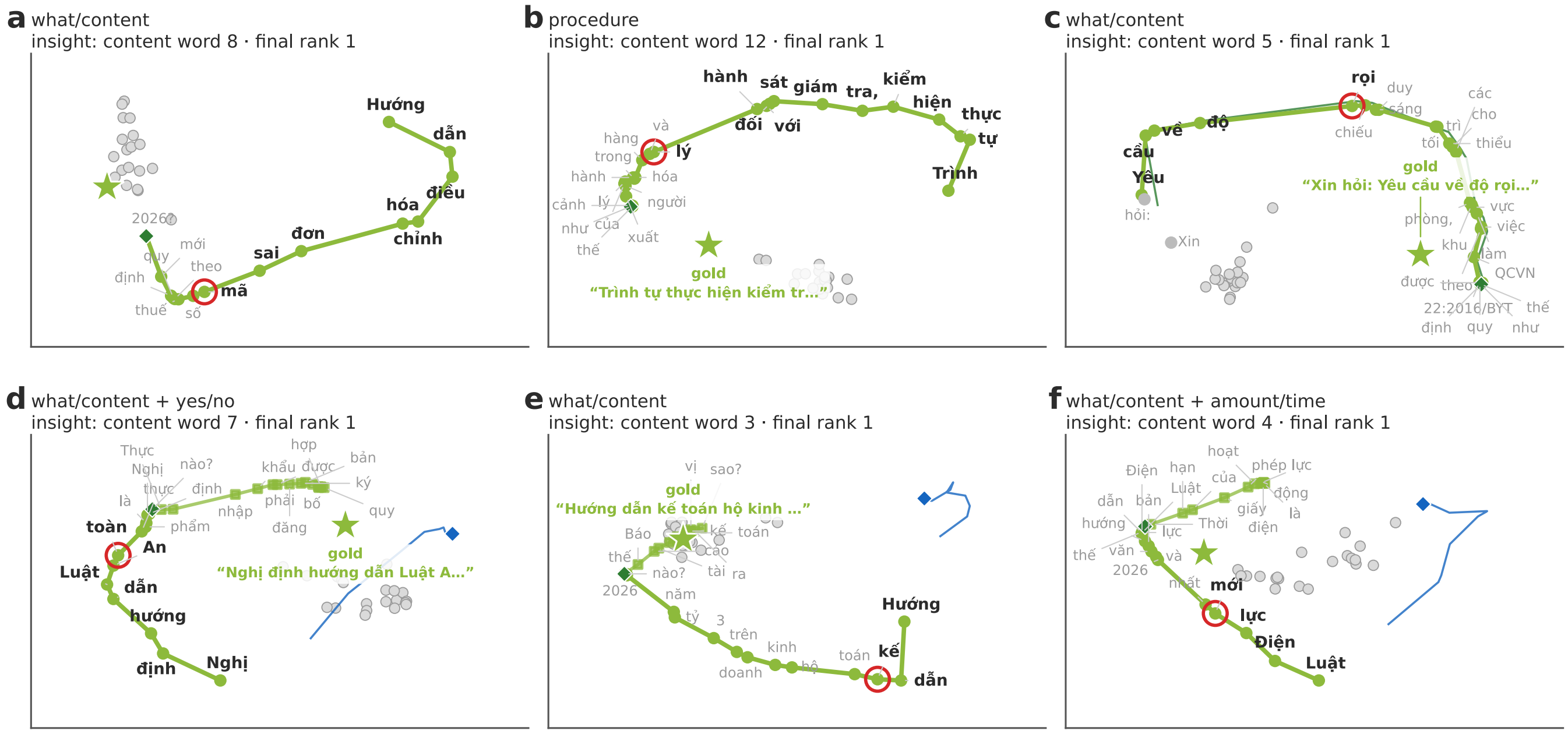

Figure 16: **Procedure headlines** (Nemotron-3-Embed-8B): (b) is a late lock—twelve content words before *Trình tự kiểm tra, giám sát …* (Procedure for inspection and supervision …) becomes specific enough.

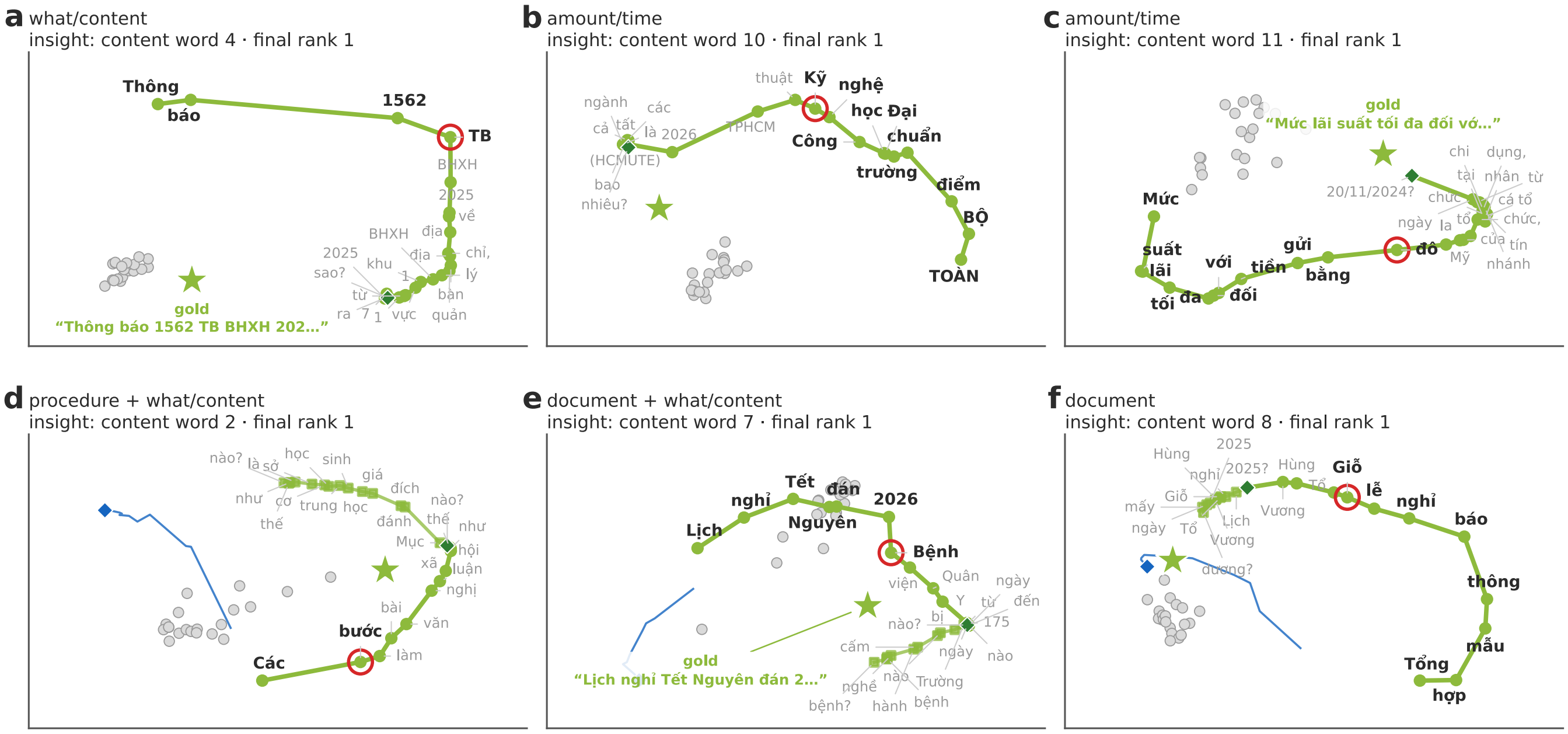

Figure 17: **Non-legal utility headlines** (Nemotron-3-Embed-8B): dates and lookup tables; the number is the lock.

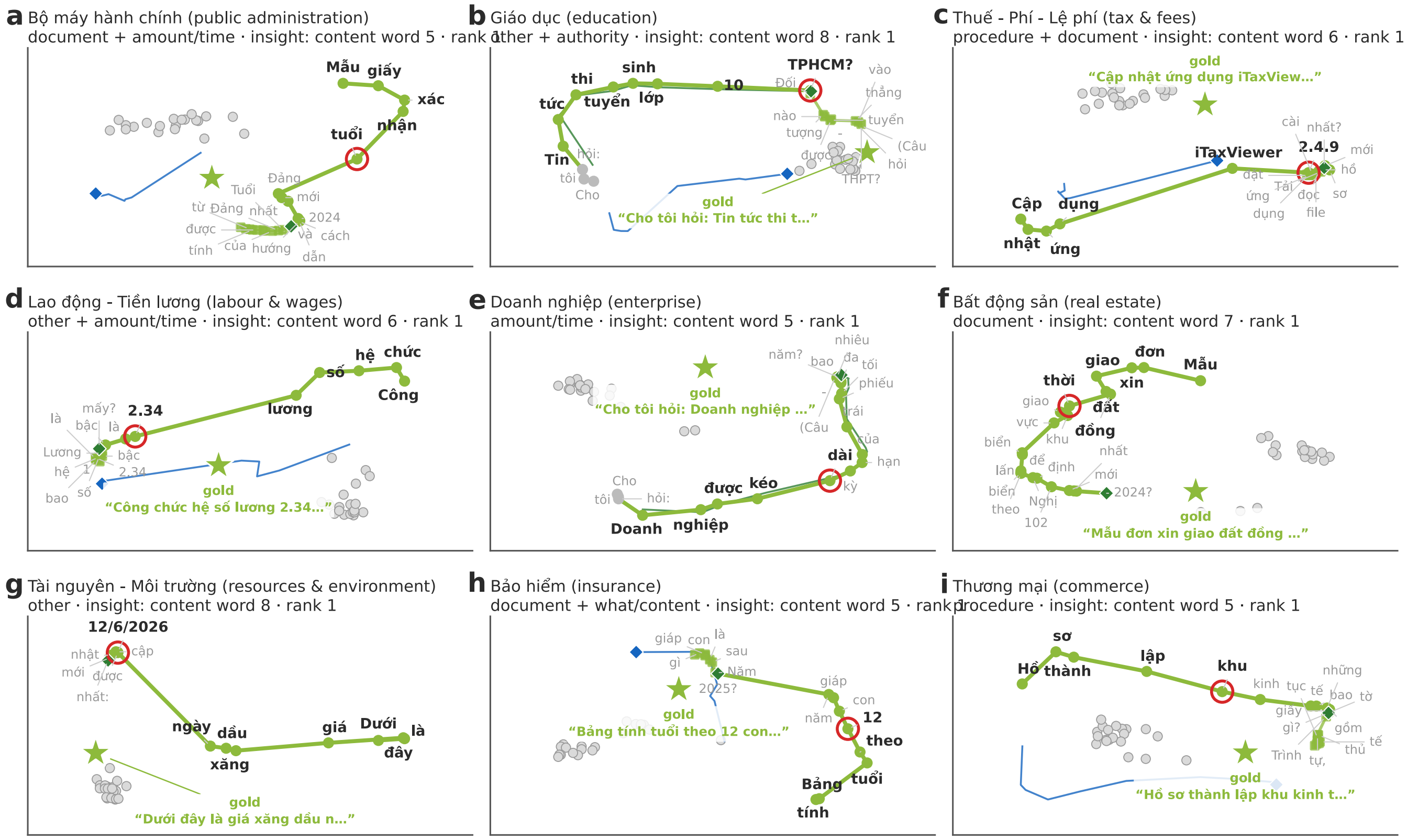

Figure 18: **One typical trajectory per legal area, areas 1–9 by size** (Nemotron-3-Embed-8B). For each area the headline whose insight point is closest to the area's median was chosen among 12–26-word headlines.

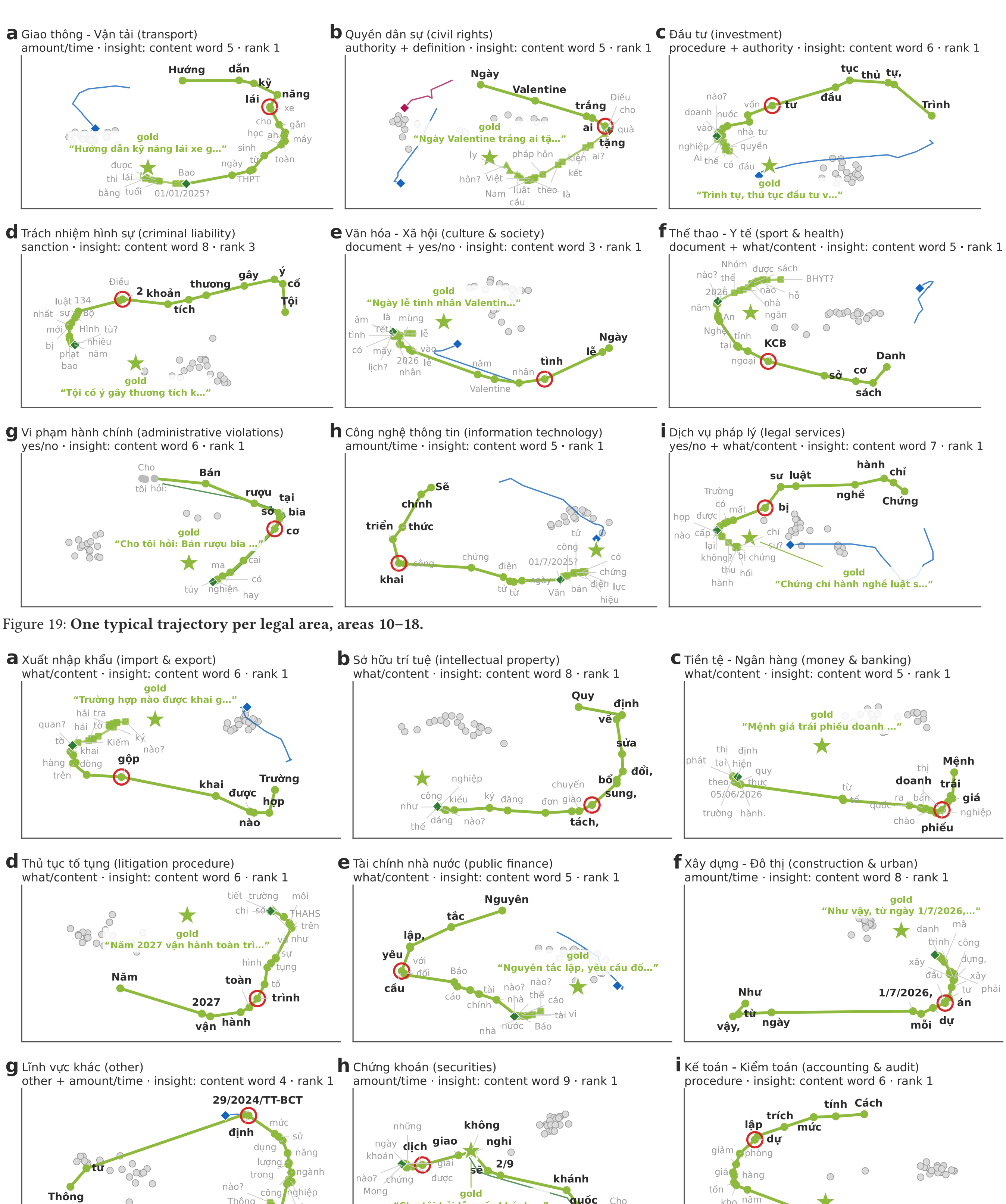

Figure 19: **One typical trajectory per legal area, areas 10–18.**

Figure 20: **One typical trajectory per legal area, areas 19–27.**